\documentclass[10pt,twocolumn,letterpaper]{article}

\usepackage{cvpr}              

\usepackage{graphicx}
\usepackage{amsmath}
\usepackage{amssymb}
\usepackage{booktabs}
\usepackage[accsupp]{axessibility}

\usepackage{multirow}

\usepackage[pagebackref,breaklinks,colorlinks]{hyperref}

\usepackage[capitalize]{cleveref}
\crefname{section}{Sec.}{Secs.}
\Crefname{section}{Section}{Sections}
\Crefname{table}{Table}{Tables}
\crefname{table}{Tab.}{Tabs.}

\def\cvprPaperID{36} 
\def\confName{CVPR}
\def\confYear{2023}

\begin{document}

\title{Benchmarking Robustness to Text-Guided Corruptions}

\author{Mohammadreza Mofayezi\thanks{indicates equal contribution} , Yasamin Medghalchi\footnotemark[1]\\
Sharif University of Technology\\
}
\maketitle

\begin{abstract}
    This study investigates the robustness of image classifiers to text-guided corruptions. 
    We utilize diffusion models to edit images to different domains.
    Unlike other works that use synthetic or hand-picked data for benchmarking, we use diffusion models as they are generative models capable of learning to edit images while preserving their semantic content. Thus, the corruptions will be more realistic and the comparison will be more informative.
    Also, there is no need for manual labeling and we can create large-scale benchmarks with less effort.
    We define a prompt hierarchy based on the original ImageNet hierarchy to apply edits in different domains.
    As well as introducing a new benchmark we try to investigate the robustness of different vision models.
    The results of this study demonstrate that the performance of image classifiers decreases significantly in different language-based corruptions and edit domains. 
    We also observe that convolutional models are more robust than transformer architectures.
    Additionally, we see that common data augmentation techniques can improve the performance on both the original data and the edited images.
    The findings of this research can help improve the design of image classifiers and contribute to the development of more robust machine learning systems. 
    The code for generating the benchmark is available at \href{https://github.com/ckoorosh/RobuText}{https://github.com/ckoorosh/RobuText}.
\end{abstract}

\section{Introduction}
\label{sec:intro}
Image classifiers are widely used in various applications such as object recognition, medical diagnosis, and autonomous driving. These systems are designed to classify images accurately and reliably, which requires them to be robust to various sources of noise and corruption \cite{recht2018cifar, hendrycks2019robustness, azulay2018deep}. 

However, recent studies have shown that image classifiers can be vulnerable to small corruptions, where subtle changes to an image can significantly impact the classifier's performance \cite{hendrycks2019robustness}. This issue has raised concerns about the robustness of image classifiers and their reliability in real-world scenarios.
Also, current benchmarks and training datasets do not cover all possible real-world situations such as weather changes, color and texture variation, or context changes \cite{zhao22oodcv}. 

In this paper, we investigate the robustness of image classifiers to text-guided corruptions using diffusion models to create image edits. Diffusion models are a type of generative model that can learn to edit images while preserving their semantic content. We use these models to create various text-guided corruptions in different domains and evaluate the performance of different image classifiers under these corruptions.

\begin{figure}
    \centering
    \includegraphics[width=1\columnwidth]{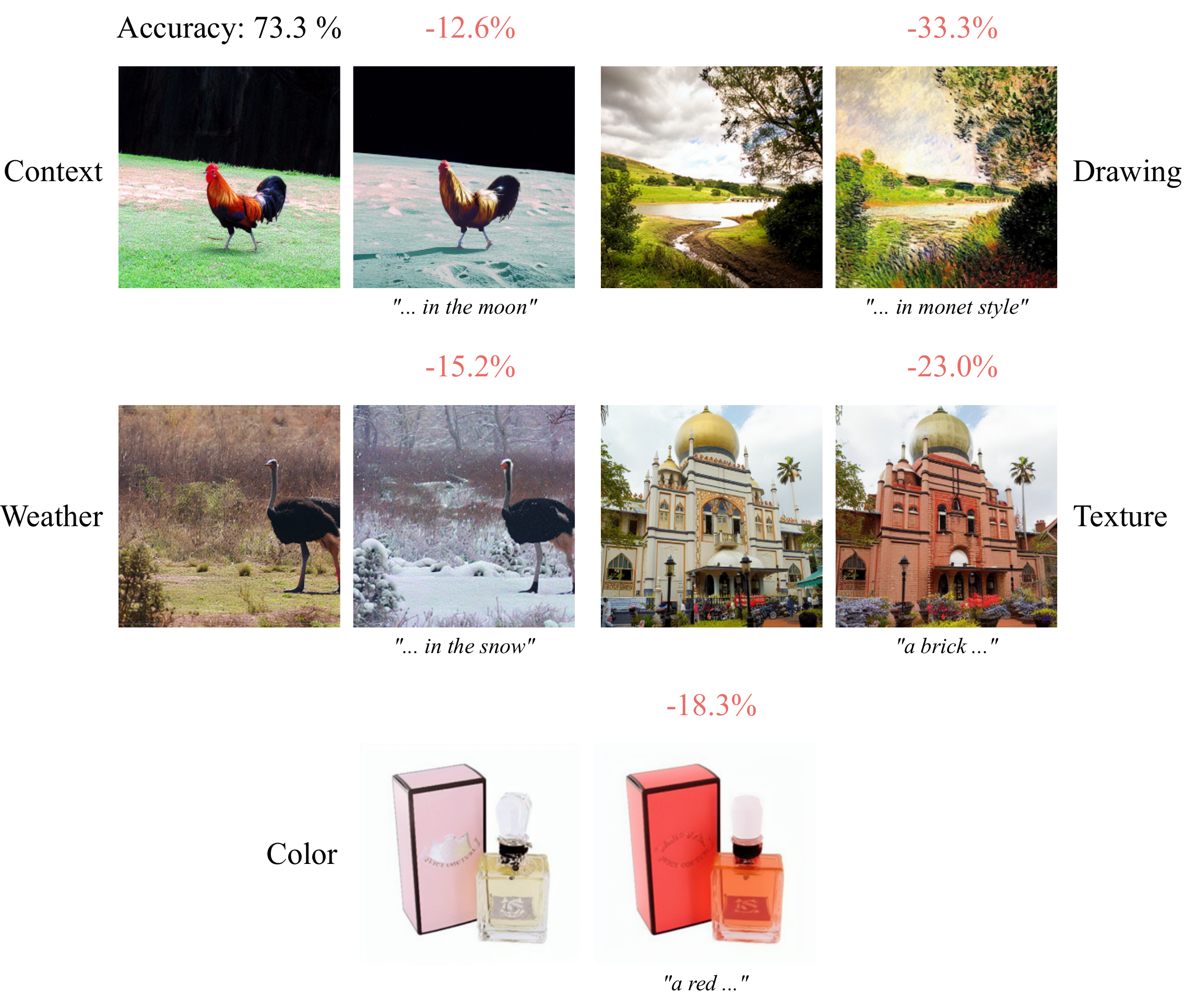}
    \caption{Current image classifiers are not robust to domain changes. We can see that when we evaluate these models on edited images the performance drops significantly. For this, we introduce a new benchmark that includes text-guided edited images in 5 different domains to assess and improve models' performance.}
    \label{fig:overview}
\end{figure}

The main objective of this study is to conduct a comprehensive benchmark of the performance of image classifiers under various language-based corruptions. 
We evaluate the classifiers' performance in different domains. 
The results of this study will provide insights into the robustness of image classifiers to language-based corruptions and identify the limitations of existing image classification systems.
We observe that convolutional models are more robust than transformers in different domains. We can also see that increasing the complexity usually improves robustness. Moreover, we check the effect of the different domains of prompts for editing on the robustness of classifiers. The results show that their accuracies fall more significantly in the drawing domain in comparison to others. Figuring out which domain of prompts is tough for handling for classifiers can give the hint for designing a new augmentation technique based on language-based models.

The remainder of this paper is organized as follows. Section 2 reviews related work in the field of image classification and robustness to corruption. Section 3 describes the methodology used to create the benchmark in this study, including the dataset and the prompts hierarchy used to create language-based corruptions and the image classifiers evaluated. Section 4 presents the experimental results and analyses the performance of different image classifiers under various language-based corruptions. Finally, Section 5 concludes the paper, and Section 6 discusses future directions for research in this field. 

Our contributions are in summary: 1) Creating a novel text-guided benchmark for evaluating the robustness of different vision models. 2) Using five domains as a testbed to evaluate image classifier performance, to gain a better understanding of where the models struggle most, and 3) Studying the effect of two common data augmentation techniques on robustness in different domains. 

\section{Related Work}
\noindent \textbf{Diffusion Models.}~~
Diffusion models are new generative models which produce images based on a forward and reverse process. In the forward process, we add a gaussian noise with certain variance and zero mean to the image in each step. The noisy image in step $t$, $x_{t}$, is produced as follow:
\begin{equation}
    x_{t} = \sqrt{\alpha_{t}} x_{0} + \sqrt{ 1 - \alpha_{t}} \epsilon,\quad 
    \epsilon ~ \sim \mathcal{N}( 0 , 1).
\end{equation}
$x_{0}$ is a clean image and $\epsilon$ is an additive noise. $\alpha_{t}$ is defined as :
\begin{equation}
        \alpha_{t} = \prod_{s=1}^{t} 1-\beta_{s}.
\end{equation}
$\beta_{t} $ can be a learned or fixed variance scheduler.
In the training procedure, the model tries to predict the noise added in the forward process.
In the reverse process, the predicted noise is subtracted from the noisy image in each step \cite{DDPM}.

\noindent \textbf{Null-text Inversion.}~~
Null-text Inversion model \cite{mokady2022null} is a text-guided diffusion model that can edit and manipulate an image using text only. This model needs source and target prompts to produce an edited image. This model has two main contributions which are pivotal inversion for diffusion models and null-text optimization. 

Most inversion works use random noise vectors for optimization while null-text performs local optimization and uses a pivotal noise vector instead, which illustrates more efficient inversion. Since they utilize Stable Diffusion \cite{rombach2021highresolution}, they require applying classifier-free guidance which has a trade-off between editability and being a good approximation. As a result, the guidance scale is fixed to 1 as their pivot trajectory and the optimization starts around it with a guidance scale larger than 1.

Null-text is inspired by the prompt-to-prompt model \cite{hertz2022prompt}. Instead of using default embedding for null text, they optimize it for each input image that can be used for multiple editing scenarios. This process is applied for each time step, $t$, and an optimized result of the previous step, $t+1$, is used as a starting point
\cite{mokady2022null}.

\noindent \textbf{Benchmarking Robustness.}~~
Recent work focuses on two approaches for creating robustness benchmarks: \textit{Synthetic} data which means utilizing synthetic images to test the robustness of neural networks and \textit{Hand-collected} data which means focusing on collecting real-world images.

Recently, there has been considerable effort devoted to testing the robustness of neural networks by employing synthetic images. One such example is ImageNet-C \cite{hendrycks2019robustness}, which assesses how neural networks perform on images with synthetic noise such as Gaussian noise, motion blur, zoom blur, and JPEG compression. To accomplish this, the standard ImageNet \cite{deng2009imagenet} validation set is modified with these noises. In addition to perturbing images with these processing pipeline noises, other research \cite{geirhosimagenet} has evaluated the shape and texture bias of deep neural networks by using images with artificially overwritten textures. Augmenting images with style transfer \cite{gatys2016image, geirhosimagenet} or by taking a linear combination of augmented images and the original images \cite{hendrycksaugmix} has been found to be a beneficial way of enhancing the robustness to these synthetic image noises or texture alterations. Nevertheless, these evaluations are limited because synthetic image perturbations are unable to accurately replicate real-world scenarios. \cite{zhao22oodcv} demonstrates that style transfer and strong augmentation are not helpful in dealing with changes in shape and pose.

Other recent studies have focused on using real-world images to evaluate the performance of deep neural network (DNN) models, as distribution shifts in these images can have a significant impact on the robustness of these models. For example, ImageNet-V2 \cite{recht2019imagenet}, which gathered images from Flickr to create a new test set for ImageNet \cite{deng2009imagenet}, found that DNN model performance decreased when faced with these real-world images. Similarly, ImageNet-A \cite{Hendrycks_2021_CVPR} utilized an adversarial filtration technique to collect a new test set of images using ResNet-50 \cite{he2016deep} that caused the performance to drop significantly when transferred to other architectures. However, ImageNet-A \cite{Hendrycks_2021_CVPR} was unable to isolate the specific factors causing the decrease in performance. More recently, ImageNet-R \cite{hendrycks2021many} collected four out-of-distribution testing benchmarks that incorporated shifts in texture, geo-location, camera parameters, and blur, and found that no single technique could improve model performance across all these factors. Other benchmarks have also been developed to evaluate how well DNN models can learn invariant features from unbalanced datasets \cite{tang2022invariant}, as well as benchmarks composed of real-world image shifts \cite{koh2021wilds}.

\section{Text-guided Robustness Benchmark}
We utilize the null-text inversion \cite{mokady2022null} model to edit ImageNet \cite{deng2009imagenet} images with our prompts. 
Our text-guided benchmark is constructed by edited images in different domains. These domains are Drawing, Weather, Color, Texture, and Context.

We will introduce the prompts hierarchy of each domain in the next section. 

\subsection{Dataset}
ImageNet \cite{deng2009imagenet} is one of the largest benchmarks for image classification. The ImageNet-1K dataset consists of 1000 object classes with 50K validation images. 
The classes are based on a hierarchical structure provided by WordNet.
Each meaningful word or phrase in Wordnet, called "synset", has about a thousand images on average in the ImageNet dataset. 
For the benchmark, the validation set of the ILSVRC-2012 dataset is used.

\subsection{ImageNet Hierarchy}
First, we needed to divide the ImageNet classes into more general sub-classes and define prompts for each subclass differently to generate more meaningful images. For instance, changing a texture to metal makes sense for tools not for animals. The super-class and sub-classes are shown in Table \ref{tab:sub-classes}.

\begin{table}[h!]
\centering
    \begin{tabular}{@{}ccc@{}}
    \toprule
    Super Class & Sub-class & Index \\ \midrule
    \multirow{3}{*}{Organism} & Animal & 1 \\
     & Plant & 2 \\
     & Person & 3 \\ \midrule
    \multirow{4}{*}{Artifact} & Vehicle & 4 \\
     & Furniture & 5 \\
     & Tool & 6 \\
     & Food & 7 \\ \midrule
    \multirow{2}{*}{Geological Formation} & Structure & 8 \\
     & Landscape & 9 \\ \bottomrule
    \end{tabular}
\caption{The ImageNet \cite{deng2009imagenet} hierarchy used for defining the sub-classes.}
\label{tab:sub-classes}
\end{table}

\subsection{Prompt Hierarchy}
One of the main problems in recent text-guided models is that all target prompts cannot be applied to whole images; so, it is necessary to find prompts that have good results on most of the images in that specific subclass.
To overcome this problem, we introduce a set of hand-engineered prompts for applying the edits to the images. To have a sight of what happened to each image and be able to analyze the impact of each category of prompts, we define 5 domains for editing prompts. It is noteworthy to mention that if the editing prompts can not be performed, the image doesn't convert to any damaged or different image. This point is vital for using this process without prompt engineering that can assure users that the originality of images is kept safe.
The full prompt hierarchy is shown in Table \ref{tab:prompts}.

\begin{table}[h!]
\centering
    \resizebox{\columnwidth}{!}{
    \begin{tabular}{@{}clc@{}}
    \toprule
    Domain & \multicolumn{1}{c}{Prompt} & Sub-class \\ \midrule
    \multirow{6}{*}{Drawing} & A watercolor painting of a {[}class{]} & 1-2-3-4-5-6-7-8-9 \\
     & An antique sketch of a {[}class{]} & 1-2-3-4-5-6-7-8-9 \\
     & A pencil drawing of a {[}class{]} & 1-2-3-4-5-6-7-8-9 \\
     & A sketch with crayon of a {[}class{]} & 1-2-3-4-5-6-7-8-9 \\
     & A {[}class{]} in Monet style & 1-2-3-4-5-6-7-8-9 \\
     & A {[}class{]} in starry night style & 1-2-3-4-5-6-7-8-9 \\ \midrule
    \multirow{4}{*}{Weather} & A {[}class{]} in the snow & 1-2-3-4-5-6-7-8-9 \\
     & A {[}class{]} on ice & 1 \\
     & A {[}class{]} in the fog & 1-2-3-4-5-6-7-8-9 \\
     & A {[}class{]} in the rain & 1-2-3-4-5-6-7-8-9 \\ \midrule
    Color & A {[}color name{]} {[}class{]} & 1-2-3-4-5-6-7-8-9 \\ \midrule
    \multirow{5}{*}{Texture} & A metal {[}class{]} & 5-6 \\
     & A wooden {[}class{]} & 5-6-8 \\
     & A glass {[}class{]} & 8 \\
     & A brick {[}class{]} & 8 \\
     & A golden {[}class{]} & 8 \\ \midrule
    \multirow{6}{*}{Context} & A {[}class{]} on the water & 1-2 \\
     & A {[}class{]} on the moon & 1-2-3-4-6 \\
     & A {[}class{]} on the mars & 1-2-3-4-6 \\
     & A {[}class{]} in the desert & 1-2-3-4-6 \\
     & A {[}class{]} on the cloth & 2-7 \\
     & A {[}class{]} on the glass & 2-7 \\ \bottomrule
    \end{tabular}
    }
\caption{The prompt hierarchy for editing images in the five mentioned domains. In each of the five domains, we introduce specific prompts for different subclasses.}
\label{tab:prompts}
\end{table}

\section{Experiments}
We feed the edited images to multiple classifiers to calculate their error in finding the correct classes. Since we want to have an interpretation of which model is more robust or to which prompt they are sensitive, we get the results of each classifier on some random images of each super-class with a random prompt of each class of prompts.

\begin{table*}[!t]
\centering
    \begin{tabular}{@{}cccccccc@{}}
    \toprule
    Model & Original & Color & Context & Drawing & Weather & Texture \\ \midrule
    AlexNet \cite{alexnet} & 55.6 & 37.9 & 44.1 & 27.4 & 40.3 & 37.1 \\
    SqueezeNet v1.0 \cite{SqueezeNet} & 58.0 & 38.8 & 46.3 & 26.6 & 43.8 & 34.9 \\
    SqueezeNet v1.1 \cite{SqueezeNet} & 58.5 & 40.8 & 45.7 & 27.0 & 44.5 & 37.1 \\ \midrule
    VGG-11 \cite{VGG} & 66.2 & 48.2 & 53.6 & 31.5 & 51.3 & 46.6 \\
    VGG-19 \cite{VGG} & 70.0 & 52.3 & 55.7 & 34.6 & 53.2 & 47.7 \\
    VGG-19+BN \cite{VGG} & 70.2 & 53.9 & 58.9 & 38.2 & 56.5 & 50.9 \\ \midrule
    DenseNet-121 \cite{densenet} & 72.1 & 53.9 & 62.0 & 41.9 & 58.4 & 51.4 \\
    DenseNet-169 \cite{densenet} & 73.6 & 56.3 & 63.1 & 44.3 & 57.7 & 52.0 \\
    DenseNet-201 \cite{densenet} & 75.2 & 57.5 & 63.7 & 44.7 & 60.2 & 52.6 \\ \midrule
    ResNet-18 \cite{resenet} & 67.9 & 51.5 & 56.5 & 40.7 & 54.1 & 46.9 \\
    ResNet-34 \cite{resenet} & 72.1 & 52.9 & 59.1 & 40.7 & 57.4 & 50.9 \\
    ResNet-50 \cite{resenet} & 73.3 & 55.0 & 60.7 & 40.0 & 58.1 & 50.3 \\
    ResNet-101 \cite{resenet} & 79.5 & 61.1 & 69.7 & 50.0 & 66.9 & 61.1 \\
    ResNet-152 \cite{resenet} & 81.4 & 63.5 & 70.0 & 49.8 & 68.8 & 62.3 \\ \midrule
    ResNeXt-50 \cite{ResNext} & 80.1 & 62.6 & 67.9 & 48.6 & 64.2 & 60.3 \\
    ResNeXt-101 \cite{ResNext} & 80.1 & 64.9 & 70.9 & 53.2 & 67.0 & 61.7 \\
    ResNeXt-101 64x4d \cite{ResNext} & 81.2 & 65.7 & 70.5 & 52.0 & 68.2 & 64.9 \\ \midrule
    ViT-B/16 \cite{VIT} & 80.2 & 61.1 & 69.3 & 50.1 & 67.4 & 56.6 \\
    ViT-B/32 \cite{VIT} & 74.1 & 57.8 & 64.5 & 48.3 & 60.6 & 55.1 \\
    ViT-L/16 \cite{VIT} & 78.7 & 60.9 & 64.3 & 47.9 & 62.0 & 55.7 \\
    ViT-L/32 \cite{VIT} & 77.3 & 60.6 & 66.0 & 50.2 & 62.1 & 56.6 \\ \midrule
    ConvNeXt-B \cite{convnext} & 82.1 & 65.7 & 72.4 & 50.8 & 69.3 & 63.1 \\ \midrule
    Swin-B \cite{swin-transformer} & 82.5 & 63.7 & 66.5 & 47.7 & 66.9 & 60.3 \\
    Swin-B v2 \cite{swin-transformer} & 82.6 & 64.5 & 69.3 & 47.7 & 67.9 & 58.9 \\ \bottomrule
    \end{tabular}%
\caption{Performance of different vision models for image classification on different domains. We report the top-1 accuracy on the original data and the edited images in different domains.}
\label{tab:data-100-10}
\end{table*}

\noindent \textbf{Experimental Setup.}~~
We pick some classes from each super-class mentioned in Table \ref{tab:sub-classes} and for each class, we choose 10 images randomly. In total, we have 1000 images for this experiment. Since we select them randomly, we can assume that their results are a good representative of a whole test set. We pick a randomly suitable prompt from each domain mentioned in Table \ref{tab:prompts} for each image.

\noindent \textbf{Baselines.}~~
Different image classifiers used in this study are based on convolutional, and transformer layers. In the following, we intend to compare the results of AlexNet \cite{alexnet}, SqueezeNet \cite{SqueezeNet}, VGG \cite{VGG}, DenseNet \cite{densenet}, ResNet \cite{resenet}, ResNeXt \cite{ResNext}, ViT \cite{VIT}, ConvNeXt \cite{convnext}, and Swin-Transformer \cite{swin-transformer}.
For SqueezeNet \cite{SqueezeNet}, we use both version 1.0 and 1.1  of the model. (SqueezeNet v1.1 model requires 2.4x less computation than SqueezeNet v1.0 without diminishing accuracy.)
For VGGNets \cite{VGG}, we chose VGG-11, VGG-19, and VGG-19 with batch normalization.
For ResNeXt \cite{ResNext}, we evaluate models with 50 and 101 layers and also the model with increased cardinality and width named ResNeXt-101 64x4d.
And finally, we chose the base and large models of ViT \cite{VIT} and the base model for ConvNeXt \cite{convnext}, and Swin-Transformer \cite{swin-transformer}.
We evaluate the performance of classifiers using the Top-1 Accuracy.

\subsection{Effect of Architecture on Robustness}
Although the Swin-Transformer models \cite{swin-transformer} have a better performance on the original data, the ConvNeXt \cite{convnext} model, the ResNeXt \cite{ResNext} models, and deep ResNet \cite{resenet} models perform better on edited images.

In VGGNets \cite{VGG}, we can see that increasing the layers results in more accuracy, and using batch normalization makes the model more robust.
In ResNets \cite{resenet}, increasing the layers makes the model more accurate and also more robust to different domains.
In ResNeXts \cite{ResNext}, the model with increased cardinality and width (ResNeXt-101 64x4d) is more robust to texture and weather edits.
Vision Transformers \cite{VIT} have competitive accuracy on original data but they are less robust than other architectures, especially in color and texture changes.


\begin{table*}[]
\centering
    \begin{tabular}{@{}ccccccc@{}}
    \toprule
    Model & Original & Color & Context & Drawing & Weather & Texture \\ \midrule
    ResNet-50 \cite{resenet} & 73.3 & 55.0 & 60.7 & 40.0 & 58.1 & 50.3 \\
    ResNet-50 \cite{resenet} + Style Transfer \cite{geirhosimagenet} & 71.5 & 55.4 & 62.9 & 43.6 & 60.1 & 51.7 \\
    ResNet-50 \cite{resenet} + AugMix \cite{hendrycksaugmix} & 75.6 & 57.7 & 64.5 & 48 & 62.4 & 55.1 \\ \bottomrule
    \end{tabular}%
\caption{Effect of data augmentation techniques on robustness. We can see that the style transfer method \cite{geirhosimagenet} has improved robustness in the drawing domain more significantly. AugMix \cite{hendrycksaugmix} has improved the performance both on the original data and on the edited data.}
\label{tab:augment}
\end{table*}

\begin{figure}
    \centering
    \includegraphics[width=1\columnwidth]{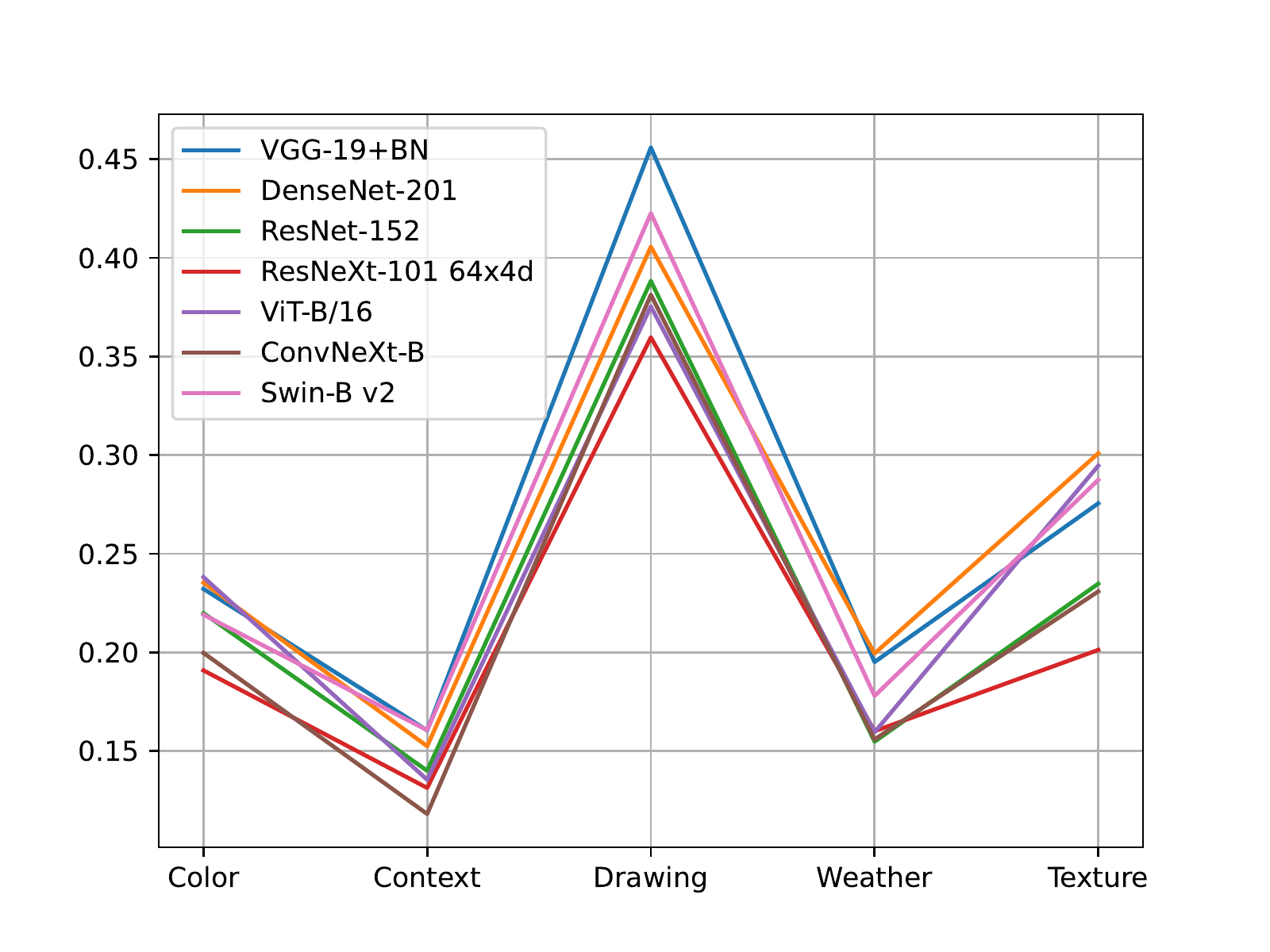}
    \caption{Robustness of various image classifiers in different domains. We can see that all models struggle in the 'Drawing' domain. We define robustness as the relative accuracy on the edited data and the original data, divided by the accuracy on the original data.}
    \label{fig:results}
\end{figure}

\subsection{Effect of Data Augmentation on Robustness}
Previous research has shown that data augmentation techniques can improve the robustness of image classifiers \cite{geirhosimagenet, hendrycksaugmix}.
Stylizing images with artistic textures \cite{geirhosimagenet} and blending the original image with a highly augmented one (known as AugMix) \cite{hendrycksaugmix} are examples of such data augmentation techniques.

We can see the effect of training a ResNet-50 \cite{resenet} with both methods in Table \ref{tab:augment}.
Using style transfer as augmentation improves the overall performance in different domains and the effect is most in the 'Drawing' domain.
AugMix \cite{hendrycksaugmix}, has a better performance in all of the domains including the original data.

\subsection{Effect of Domains on Robustness}
In this section, we aim to observe for each class what kind of manipulation is riskier in the image classification task. According to Figure \ref{fig:results}, all domains decrease the accuracy of classifiers but the drawing domain is the most difficult domain for classifiers. This interpretation is advantageous for designing a new augmentation technique based on drawing domains and also comparing the performance of various classifiers in this tough domain in the object detection task. Another interesting point is that although different models have dissimilar accuracy results, they have the same behavior about different domains, and the order of hardness of domains is approximately alike for all of them.

\subsection{Ablation Study}
We also evaluated the models' performance on the animal subclass of the ImageNet-1K dataset. The results are provided in Table \ref{tab:animal-10-10}. 
We chose 10 classes from the subclass and 10 images per class.

In ResNets \cite{resenet}, increasing the layers makes the model more accurate and robust to different domains.
Similar to Table \ref{tab:data-100-10}, we can see that batch normalization has made VGGNet more robust.

\begin{table}[]
\centering
\resizebox{\columnwidth}{!}{%
    \begin{tabular}{@{}cccccc@{}}
    \toprule
    Model & Original & Color & Context & Drawing & Weather \\ \midrule
    AlexNet \cite{alexnet} & 60 & 40 & 42 & 26 & 51 \\
    SqueezeNet v1.0 \cite{SqueezeNet} & 60 & 37 & 57 & 36 & 55 \\
    SqueezeNet v1.1 \cite{SqueezeNet} & 64 & 39 & 52 & 30 & 54 \\ \midrule
    VGG-11 \cite{VGG} & 80 & 60 & 57 & 39 & 71 \\
    VGG-19 \cite{VGG} & 82 & 58 & 62 & 39 & 69 \\
    VGG-19+BN \cite{VGG} & 82 & 63 & 72 & 45 & 72 \\ \midrule
    DenseNet-121 \cite{densenet} & 85 & 59 & 71 & 45 & 70 \\
    DenseNet-169 \cite{densenet} & 87 & 58 & 77 & 52 & 77 \\
    DenseNet-201 \cite{densenet} & 83 & 62 & 71 & 47 & 67 \\ \midrule
    ResNet-18 \cite{resenet} & 76 & 52 & 63 & 42 & 65 \\
    ResNet-34 \cite{resenet} & 78 & 52 & 65 & 41 & 67 \\
    ResNet-50 \cite{resenet} & 85 & 60 & 68 & 50 & 68 \\
    ResNet-101 \cite{resenet} & 88 & 70 & 81 & 63 & 75 \\
    ResNet-152 \cite{resenet} & 89 & 72 & 78 & 63 & 79 \\ \midrule
    ResNeXt-50 \cite{ResNext} & 88 & 68 & 79 & 56 & 78 \\
    ResNeXt-101 \cite{ResNext} & 88 & 79 & 79 & 64 & 82 \\
    ResNeXt-101 64x4d \cite{ResNext} & 92 & 76 & 78 & 64 & 84 \\ \midrule
    ViT-B/16 \cite{VIT} & 83 & 68 & 76 & 52 & 69 \\
    ViT-B/32 \cite{VIT} & 81 & 62 & 68 & 53 & 65 \\
    ViT-L/16 \cite{VIT} & 89 & 65 & 79 & 57 & 72 \\
    ViT-L/32 \cite{VIT} & 81 & 69 & 74 & 57 & 69 \\ \midrule
    ConvNeXt-B \cite{convnext} & 88 & 76 & 82 & 64 & 81 \\ \midrule
    Swin-B \cite{swin-transformer} & 90 & 72 & 81 & 53 & 75 \\
    Swin-B v2 \cite{swin-transformer} & 90 & 77 & 80 & 56 & 79 \\ \bottomrule
    \end{tabular}%
    }
\caption{Performance of different vision models for image classification on the animal subset. We report the top-1 accuracy on the original data and the edited images in different domains.}
\label{tab:animal-10-10}
\end{table}

\begin{figure*}
    \centering
    \includegraphics[width=1\textwidth]{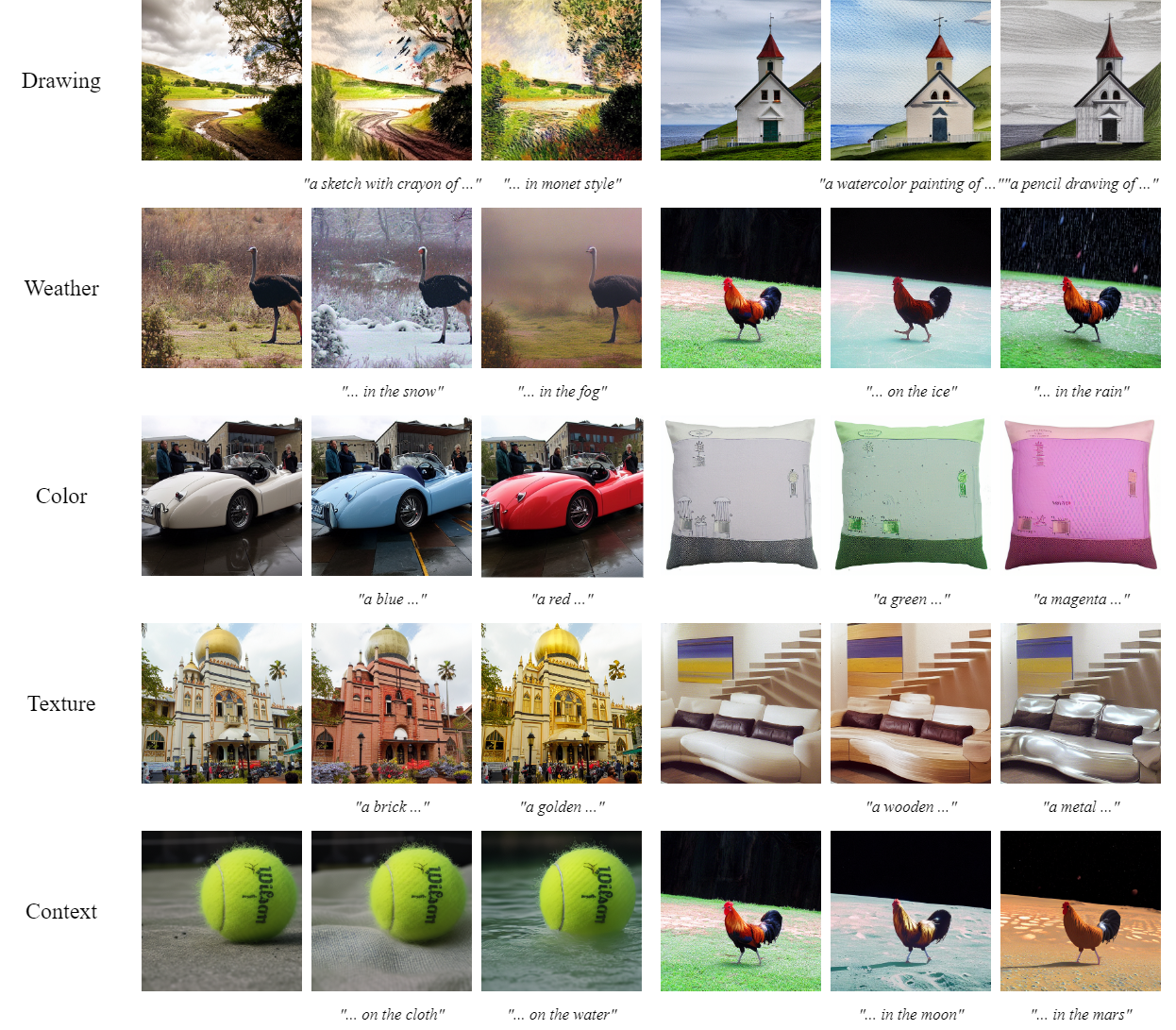}
    \caption{Visualization of edited images with different prompts in the drawing, weather, color, texture, and context domains. With our benchmark vision models can be further evaluated with more realistic and diverse data.}
    \label{fig:visual}
\end{figure*}

\section{Conclusion}
In this study, we provide an automated process to edit images with language-based diffusion models which have a great performance in generating images. First, we divide the ImageNet classes into some super classes based on the WordNet hierarchy and we assigned some prompts for each subclass in 5 domains. Second, we compare the robustness of different classification baselines with this newly edited benchmark, which illustrates that convolutional models are more robust. Furthermore, we compare the results of different domains of prompts and figure out that the drawing domain is the hardest editing domain for networks to handle. As our final step, we can mention that the data augmentation technique can improve the performance on both the new corrupted and the original images. This study can be a first step for benchmarking different test sets in a fully automated way for any dataset. Moreover, it can give an impressive insight into the weakness of models in diverse domains.

\section{Future Work}
This manipulation procedure with text-guided models can be added to the training process as a new automated augmentation technique. Since previous research \cite{hendrycksaugmix, geirhosimagenet} shows the effectiveness of augmented techniques, we expect that augmentation with language-based models can outperform the previous methods. We highly recommend optimizing prompts for each image as an adversarial attack on the network. As a result, we can assign the best prompts for each image in a fully automated way to fool the model.

{\small
\bibliographystyle{ieee_fullname}
\bibliography{egbib}

\begin{thebibliography}{10}\itemsep=-1pt

\bibitem{azulay2018deep}
Aharon Azulay and Yair Weiss.
\newblock Why do deep convolutional networks generalize so poorly to small
  image transformations?
\newblock {\em arXiv preprint arXiv:1805.12177}, 2018.

\bibitem{deng2009imagenet}
Jia Deng, Wei Dong, Richard Socher, Li-Jia Li, Kai Li, and Li Fei-Fei.
\newblock Imagenet: A large-scale hierarchical image database.
\newblock In {\em 2009 IEEE Conference on Computer Vision and Pattern
  Recognition}, pages 248--255, 2009.

\bibitem{VIT}
Alexey Dosovitskiy, Lucas Beyer, Alexander Kolesnikov, Dirk Weissenborn,
  Xiaohua Zhai, Thomas Unterthiner, Mostafa Dehghani, Matthias Minderer, Georg
  Heigold, Sylvain Gelly, Jakob Uszkoreit, and Neil Houlsby.
\newblock An image is worth 16x16 words: Transformers for image recognition at
  scale.
\newblock {\em ICLR}, 2021.

\bibitem{gatys2016image}
Leon~A Gatys, Alexander~S Ecker, and Matthias Bethge.
\newblock Image style transfer using convolutional neural networks.
\newblock In {\em Proceedings of the IEEE Conference on Computer Vision and
  Pattern Recognition}, pages 2414--2423, 2016.

\bibitem{geirhosimagenet}
Robert Geirhos, Patricia Rubisch, Claudio Michaelis, Matthias Bethge, Felix~A
  Wichmann, and Wieland Brendel.
\newblock Imagenet-trained cnns are biased towards texture; increasing shape
  bias improves accuracy and robustness.
\newblock In {\em International Conference on Learning Representations}, 2018.

\bibitem{he2016deep}
Kaiming He, Xiangyu Zhang, Shaoqing Ren, and Jian Sun.
\newblock Deep residual learning for image recognition.
\newblock In {\em Proceedings of the IEEE Conference on Computer Vision and
  Pattern Recognition}, pages 770--778, 2016.

\bibitem{resenet}
Kaiming He, Xiangyu Zhang, Shaoqing Ren, and Jian Sun.
\newblock Deep residual learning for image recognition.
\newblock In {\em Proceedings of the IEEE conference on computer vision and
  pattern recognition}, pages 770--778, 2016.

\bibitem{hendrycks2021many}
Dan Hendrycks, Steven Basart, Norman Mu, Saurav Kadavath, Frank Wang, Evan
  Dorundo, Rahul Desai, Tyler Zhu, Samyak Parajuli, Mike Guo, et~al.
\newblock The many faces of robustness: A critical analysis of
  out-of-distribution generalization.
\newblock In {\em Proceedings of the IEEE/CVF International Conference on
  Computer Vision}, pages 8340--8349, 2021.

\bibitem{hendrycks2019robustness}
Dan Hendrycks and Thomas Dietterich.
\newblock Benchmarking neural network robustness to common corruptions and
  perturbations.
\newblock {\em Proceedings of the International Conference on Learning
  Representations}, 2019.

\bibitem{hendrycksaugmix}
Dan Hendrycks, Norman Mu, Ekin~Dogus Cubuk, Barret Zoph, Justin Gilmer, and
  Balaji Lakshminarayanan.
\newblock Augmix: A simple data processing method to improve robustness and
  uncertainty.
\newblock In {\em International Conference on Learning Representations}, 2019.

\bibitem{Hendrycks_2021_CVPR}
Dan Hendrycks, Kevin Zhao, Steven Basart, Jacob Steinhardt, and Dawn Song.
\newblock Natural adversarial examples.
\newblock In {\em Proceedings of the IEEE/CVF Conference on Computer Vision and
  Pattern Recognition (CVPR)}, pages 15262--15271, June 2021.

\bibitem{hertz2022prompt}
Amir Hertz, Ron Mokady, Jay Tenenbaum, Kfir Aberman, Yael Pritch, and Daniel
  Cohen-Or.
\newblock Prompt-to-prompt image editing with cross attention control.
\newblock {\em arXiv preprint arXiv:2208.01626}, 2022.

\bibitem{DDPM}
Jonathan Ho, Ajay Jain, and Pieter Abbeel.
\newblock Denoising diffusion probabilistic models.
\newblock {\em arXiv preprint arxiv:2006.11239}, 2020.

\bibitem{densenet}
Gao Huang, Zhuang Liu, Laurens van~der Maaten, and Kilian~Q Weinberger.
\newblock Densely connected convolutional networks.
\newblock In {\em Proceedings of the IEEE Conference on Computer Vision and
  Pattern Recognition}, 2017.

\bibitem{SqueezeNet}
Forrest~N. Iandola, Song Han, Matthew~W. Moskewicz, Khalid Ashraf, William~J.
  Dally, and Kurt Keutzer.
\newblock Squeezenet: Alexnet-level accuracy with 50x fewer parameters and
  $<$0.5mb model size.
\newblock {\em arXiv:1602.07360}, 2016.

\bibitem{koh2021wilds}
Pang~Wei Koh, Shiori Sagawa, Henrik Marklund, Sang~Michael Xie, Marvin Zhang,
  Akshay Balsubramani, Weihua Hu, Michihiro Yasunaga, Richard~Lanas Phillips,
  Irena Gao, et~al.
\newblock Wilds: A benchmark of in-the-wild distribution shifts.
\newblock In {\em International Conference on Machine Learning}, pages
  5637--5664. PMLR, 2021.

\bibitem{alexnet}
Alex Krizhevsky, Ilya Sutskever, and Geoffrey~E Hinton.
\newblock Imagenet classification with deep convolutional neural networks.
\newblock In {\em Advances in Neural Information Processing Systems},
  volume~25, 2012.

\bibitem{swin-transformer}
Ze Liu, Yutong Lin, Yue Cao, Han Hu, Yixuan Wei, Zheng Zhang, Stephen Lin, and
  Baining Guo.
\newblock Swin transformer: Hierarchical vision transformer using shifted
  windows.
\newblock In {\em Proceedings of the IEEE/CVF International Conference on
  Computer Vision (ICCV)}, 2021.

\bibitem{convnext}
Zhuang Liu, Hanzi Mao, Chao-Yuan Wu, Christoph Feichtenhofer, Trevor Darrell,
  and Saining Xie.
\newblock A convnet for the 2020s.
\newblock {\em Proceedings of the IEEE/CVF Conference on Computer Vision and
  Pattern Recognition (CVPR)}, 2022.

\bibitem{mokady2022null}
Ron Mokady, Amir Hertz, Kfir Aberman, Yael Pritch, and Daniel Cohen-Or.
\newblock Null-text inversion for editing real images using guided diffusion
  models.
\newblock {\em arXiv preprint arXiv:2211.09794}, 2022.

\bibitem{recht2018cifar}
Benjamin Recht, Rebecca Roelofs, Ludwig Schmidt, and Vaishaal Shankar.
\newblock Do cifar-10 classifiers generalize to cifar-10?
\newblock {\em arXiv preprint arXiv:1806.00451}, 2018.

\bibitem{recht2019imagenet}
Benjamin Recht, Rebecca Roelofs, Ludwig Schmidt, and Vaishaal Shankar.
\newblock Do imagenet classifiers generalize to imagenet?
\newblock In {\em International conference on machine learning}, pages
  5389--5400. PMLR, 2019.

\bibitem{rombach2021highresolution}
Robin Rombach, Andreas Blattmann, Dominik Lorenz, Patrick Esser, and Björn
  Ommer.
\newblock High-resolution image synthesis with latent diffusion models, 2021.

\bibitem{VGG}
Andrew~Zisserman Simonyan, Karen.
\newblock Very deep convolutional networks for large-scale image recognition.
\newblock {\em arXiv:1409.1556}, 2014.

\bibitem{tang2022invariant}
Kaihua Tang, Mingyuan Tao, Jiaxin Qi, Zhenguang Liu, and Hanwang Zhang.
\newblock Invariant feature learning for generalized long-tailed
  classification.
\newblock In {\em Proceedings of the European Conference on Computer Vision
  (ECCV)}, pages 709--726, 2022.

\bibitem{ResNext}
Saining Xie, Ross Girshick, Piotr Dollár, Zhuowen Tu, and Kaiming He.
\newblock Aggregated residual transformations for deep neural networks.
\newblock {\em arXiv preprint arXiv:1611.05431}, 2016.

\bibitem{zhao22oodcv}
Bingchen Zhao, Shaozuo Yu, Wufei Ma, Mingxin Yu, Shenxiao Mei, Angtian Wang, Ju
  He, Alan Yuille, and Adam Kortylewski.
\newblock Ood-cv: A benchmark for robustness to out-of-distribution shifts of
  individual nuisances in natural images.
\newblock {\em Proceedings of the European Conference on Computer Vision
  (ECCV)}, 2022.

\end{thebibliography}
}

\end{document}